# Improved YOLOv8 Detection Algorithm in X-ray Contraband

**Liyao Lu**  2220011071@student.must.edu.mo

*College of Innovation Engineering*

*Macau University of Science and Technology*

*Macau, 999078, China*

**Corresponding Author:** Liyao Lu.

## Abstract

Security inspection is the first line of defense to ensure the safety of people's lives and property, and intelligent security inspection is an inevitable trend in the future development of the security inspection industry. Aiming at the problems of overlapping detection objects, false detection of contraband, and missed detection in the process of X-ray image detection, an improved X-ray contraband detection algorithm CSS-YOLO based on YOLOv8s is proposed. Firstly, aim at the missed detection problem in target detection. We choose to use the convolutional block attention module (CBAM) in the backbone, where the channel attention module and spatial attention module can effectively extract the feature information flow in the network. Secondly, for the small targets object detection. Swin Transformer module is introduced into the YOLOv8s backbone network. Utilize its Self-Attention and Shifted Window to improve the ability of the model to extract global features from X-ray images. Finally, to the overlapping problem for detection targets. Use Soft NMS to avoid missing objects by removing overlapping proposals. The average detection accuracy of the algorithm in the SIXray positive sample data set is 72.3%, which is 2.5 percentage points higher than that of YOLOv8s. Aiming at the messy X-ray detection sets on the market, we collected and sorted out a self-made data set QXray that is closer to the actual situation through major data websites and carried out simulated detection on it. In the self-made data set QXray, the average detection accuracy rate reaches 72.6%, which is 12% higher than that of YOLOv8s. The experimental results show that the detection accuracy of the improved algorithm is significantly improved compared with the original YOLOv8s algorithm, which proves the effectiveness of the algorithm. Our implementation code is published at https://github.com/luliyaoLeo/Improved-YOLOv8-Detection-Algorithm-in-X-ray-Security-Inspection-Image.

# 1. INTRODUCTION

X-ray luggage screening [1-2] is widely used to maintain aviation and transportation safety, to ensure the safety of the people and prevent the occurrence of danger. Some important transportation hubs and densely populated public places use X-ray machines to conduct safety checks on passengers' luggage. How to achieve automatic identification and intelligent detection of prohibited items has high practical research significance.

The current commonly used detection method is to pass luggage through an X-ray machine and generate real-time X-ray images. Security personnel visually inspect for the presence of prohibited items. The drawback of this is that if there are many people and luggage. The visual judgment of security personnel can slow down the speed of security checks, resulting in traffic congestion. At the same time, workers who work in this situation for a long time may feel tired and lack concentration, leading to missed or false detections, seriously affecting people's travel efficiency.

With the continuous development of computer vision, the detection method of prohibited items in X-ray security inspection images based on convolutional neural networks has achieved certain research results. Researchers hope to achieve the functions of detecting, locating, and warning prohibited items by building neural networks. Achieve automated detection without the need for security personnel to perform human eye discrimination detection.

Bastan[3] used Bag-of-Visual-Words (BOVW) model to classify contraband in images. But the performance was poor because it could not accurately identify the type and location of contraband; Zhang[4] used the characteristics of X-ray images to extract the color, texture and other features in them for detection; Aiming at the problem of unbalanced contraband categories, Miao[5] designed a Class Balance Hierarchical Refinement (CHR) model to improve the learning ability of the model for overlapping contraband; in 2020, Wei[6] improved the De-occlusion attention module to enhance Network extraction of contraband edge texture features material information. Guo[1] proposed a single-stage dual-network contraband detection algorithm, and the feature expression ability of the network has been improved to a certain extent. In 2021, Zhu[7] designed a contraband detector based on a deep convolutional neural network for the small size and overlapping occlusion of contraband targets. Zhu[8] proposed an Attention-based Multi-Scale Object Detection Network(AMOD-Net) and designed a deep feature fusion structure based on this network. Aiming at the problems of missed detection and false detection of contraband, Miao[9] proposed an improved capsule network model. In this model, the detection ability of the model in complex scenes is improved by adding a feature enhancement module. In 2022, Wu[10] propose an improved YOLOv5 X-ray security image contraband target detection algorithm. The algorithm adds the mask self-attention mechanism module to enhance the expression ability of feature information. At the same time, the introduction of Quality Focal Loss function effectively alleviates the problem of category imbalance and improves the positioning accuracy of the target.

The current contraband detection methods (such as Faster RCNN, YOLO and CenterNet, etc.) are not intuitive enough, and there are three problems as follows. (1) The model does not directly predict the target frame but uses alternating regression and classification to predict the center of the detection window. (2) The effect of the model will be affected by a series of problems, such as using post-processing to remove many overlapping frames, and the design of anchor points directly affects the effect of the model. (3) In the contraband detection task, there is no unified treatment for the size of prohibited objects, overlapping occlusion and other issues. In recent years, "CNN+Transformer" has also been used for target detection [13]. Although the performance is not bad, the model convergence speed is too slow. At the same time, the self-attention of a single Transformer will lead to too low resolution of the feature map, resulting in poor detection performance of small objects. To solve the problems of complex background interference and overlapping contraband in contraband detection. We decided to use the self-attention mechanism in the Transformer model to implant YOLOv8 to enhance the model's ability to extract global features. At the same time, the convolutional block attention module [14] is added to the YOLOv8s model, and the same-channel attention mechanism and spatial attention mechanism in the CBAM module are used to make the model pay more attention to areas containing contraband. Finally, Soft NMS [15] is used to avoid missed and false detections of contraband due to overlapping detected objects.

YOLO (You Only Look Once) is a target detection network proposed by Redmon in 2015. Based on mature changes and field deployment, YOLOv5 is a commonly used deployment model in industry. In 2022, the Ultralytics team announced the source code of the YOLOv8 network. Because of its fast, accurate, and easy-to-use design concept, its detection performance exceeds many improved v5 models, but it still needs to be improved in occluded targets and small target detection. Therefore, this paper chooses to use YOLOv8 as the basic algorithm and embeds a plug-and-play module to obtain a target detection algorithm that is more suitable for X-ray security inspection images.

Our contributions can be summarized as the following three points: 1. We made targeted improvements to the problems existing in contraband detection and achieved high detection accuracy. 2. In view of the current lack of contraband data sets, collect data sets with the best positive and negative ratios through data set websites such as Kaggle and Roboflow, aiming to provide real professional detection data for subsequent research scholars.
The remainder of this paper is organized as follows. In Section II, the improved modules we use are briefly explained. Section III introduces the datasets we use and our datasets QXary. Section IV introduces the evaluation indicators used in this study. Section V presents the experimental results. Section VI concludes this paper.

## 2. Improved YOLOv8 mode

YOLOv8 algorithm architecture consists of Backbone, Neck, and Head. In the Backbone structure, YOLOv8 uses the concept of cross level components (CSP) [12]. The Neck section utilizes multi-scale feature fusion of images. And different from YOLOv5 using the coupled head, YOLOv8 uses the decoupled head instead, separating the classification and detection head. Our improvements focus on Backbone and Neck. The improved model CSS-YOLO is shown in Figure 1.

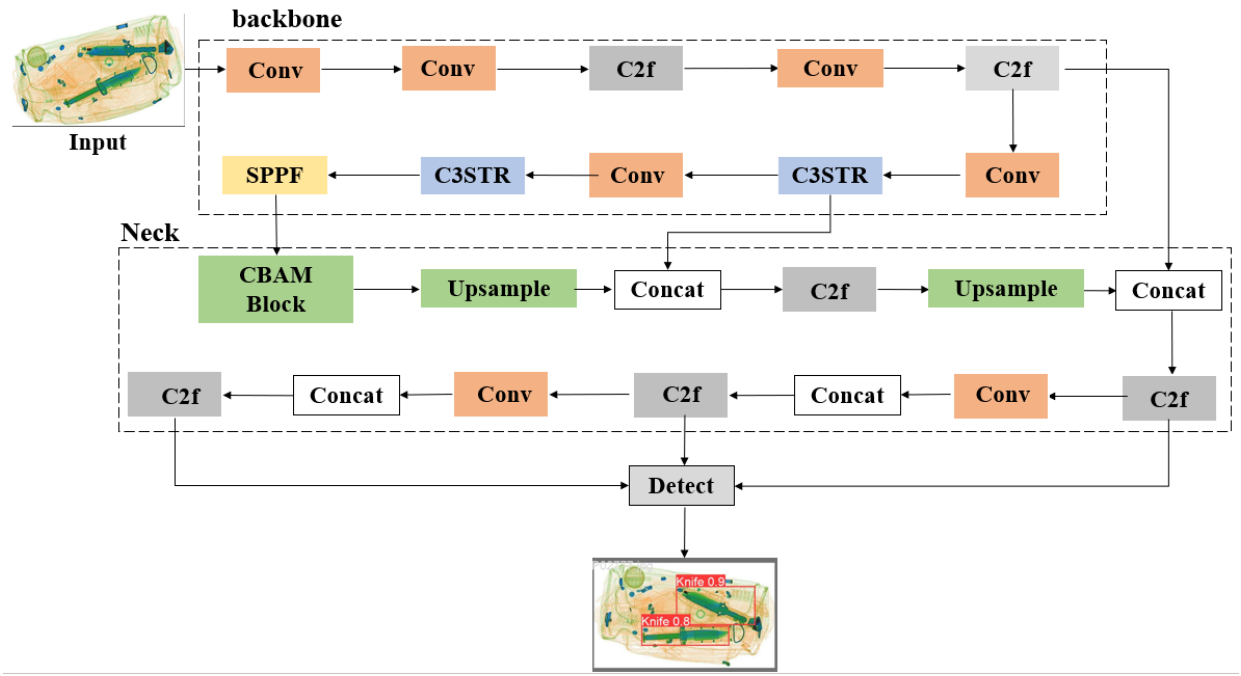

Figure 1: Improved overall structure.

## 2.1 CBAM

CBAM is an attention module for feed-forward convolutional neural networks. The principle of CBAM is to reduce the amount of computation and parameters in the model by decomposing the attention mechanism in the three-dimensional feature map into channel attention and spatial attention. Since CBAM is a lightweight general-purpose module, it can ignore module overhead and seamlessly integrate into any CNN architecture for end-to-end training. Woo[14] and others tried different attention combination sequences, and the best result was to add the spatial attention module after the channel attention module. as shown in Figure 2.

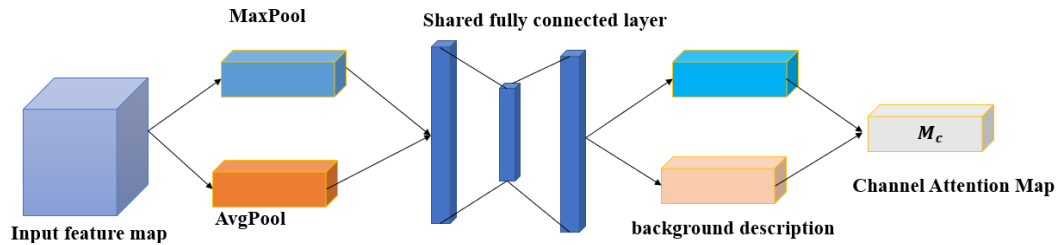

Figure 2: Channel Attention Module in CBAM

Channel attention is to compress the feature map F in the spatial dimension to obtain a one-dimensional vector. Maximum pooling and average pooling are performed within each channel. Average pooling and max pooling can be used to aggregate the spatial information of the feature maps, send them to a shared network, and compress the spatial dimensionality of the input feature maps. Combining each element by summation finally produces a channel attention map. The channel attention mechanism is shown in Equation (1), where σ denotes the sigmoid function, $W_o \in R^{C \times C/r}$, and $W_1 \in R^{C \times C/r}$. The MLP weights, $W_o$ and $W_1$, are shared for both inputs and the ReLU activation function is followed by $W_o$.

$$M_C(F) = \sigma(MLP(AvgPool(F)) + MLP(MaxPool(F)))$$
$$= \sigma(W_1(W_0(F_{avg}^c)) + W_1(W_0(F_{max}^c))) \quad (1)$$

The spatial attention module in CBAM is shown in Figure 3.

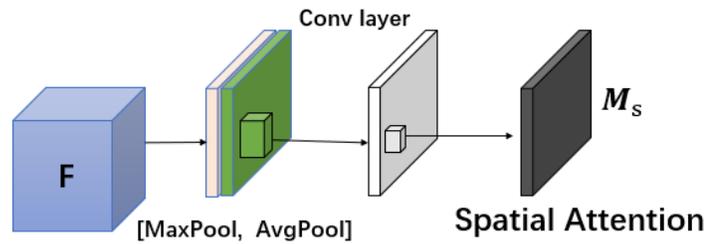

Figure 3: Spatial Attention Module in CBAM

This module multiplies the output of the channel attention module with F and takes the resulting feature map as input. The maximum pooling and average pooling between channel dimensions are used for channel compression, then the two are spliced. Finally, the weight is obtained by using 7×7 convolution and the Sigmoid function, as shown in formula (2), where σ denotes the sigmoid function and f^(7×7) represents a convolution operation with the filter size of 7 × 7.

$$M_s(F) = \sigma(f^{7 \times 7}(F_{avg}^S, F_{max}^S)) \quad (2)$$

In view of the missed detection problem in contraband detection, we decided to introduce CBAM. Use its dual-channel nature to obtain more feature information, and finally achieve the maximum detection of the quantity and location information of contraband. Through experiments, we can conclude that the model added with CBAM can better detect the missed objects. For example, the second Wrench in Figure 4 was successfully detected.

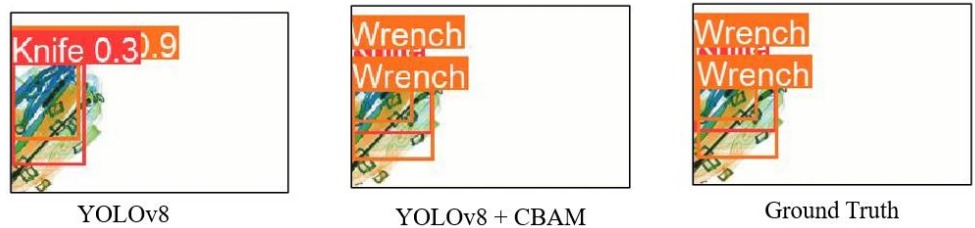

Figure 4: Test result comparison chart

## 2.2  Swin Transformer

Aiming at the size difference of contraband in X-ray images, this paper introduces the sliding window multi-head self-attention model in Swin Transformer to enhance the model's ability to extract global features. Swin Transformer is shown in Figure 5.

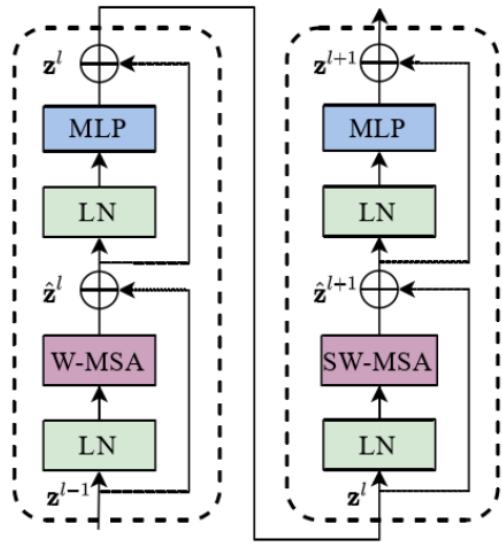

Figure 5: Swin Transformer Model

Swin Transformer consists of a window multi-head self-attention module (W-MSA) with a multi-layer perceptron and a sliding window multi-head self-attention model (shifted-window multi-head self-attention modules , SW-MSA), MSA is modeled globally, and W-MSA models through the local window. One LN layer is added before each MSA module and each MLP module, and a residual connection module is added before each MSA module and MLP module. SW-MSA shifts two patches to the lower right corner based on W-MSA, thereby indirectly expanding the receptive field, and making full use of its global features. Figure 6 is a visualization of the heatmap of the detection image. From the figure, we can see that the image using the attention mechanism will pay more attention to the features of the detected object.

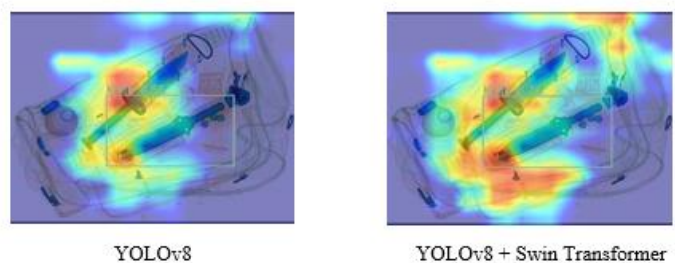

Figure 6: Swin Transformer Model

## 2.3 Soft NMS

Our paper chooses to use the Soft NMS algorithm of the Gaussian reset method to replace the NMS algorithm in the YOLOv8 algorithm. The advantage of this algorithm is that it only needs to make simple changes to the traditional NMS algorithm without adding additional parameters. Easy to modify and innovate. At the same time, Soft-NMS has the same algorithmic complexity as traditional NMS, without increasing the complexity of the algorithm. No additional training is required and easy to implement. It can be easily integrated into any object detection process. Avoid object miss detection due to removing overlapping proposals [15].

Due to the overlapping characteristics of X-ray images, the use of NMS will lead to missed detection of items in the overlapping area. Compared with the NMS algorithm that directly deletes the candidate boxes whose IOU is greater than the threshold, the Soft NMS algorithm reduces the confidence of the candidate boxes. In Soft NMS, there are two reset methods, linear and Gaussian, to reduce confidence. The linear equation is shown in equation (3).

$$S_i \begin{cases} S_i, iou(A-b_i) < N_t \\ S_i(1-iou(A-b_i)), iou(A-b_i) \geq N_t \end{cases} \quad (3)$$

Among them, s is the confidence level, and i is the serial number of the remaining boxes except for the "A" box with the highest score, sorted from high to low. $N_t$ is the specified threshold, and $b_i$ is the box to be processed. When the IOU is greater than or equal to the threshold, the confidence of the candidate box is reduced. However, the reset in a linear manner is not a continuous function, and when the IOU reaches the threshold, there may be a sudden change in the confidence. The Gaussian reset method can solve this problem, as shown in formula (4), where D represents the filtered candidate box and is a hyperparameter.

$$S_i = S_i e^{\frac{-iou(A-b_i)^2}{\sigma}}, \forall b_i \notin D \quad (4)$$

In this paper, the Soft NMS algorithm of the Gaussian reset method is used to replace the NMS algorithm in the YOLOv8 algorithm, to avoid the target missed detection caused by the deletion of overlapping candidate frames. According to the experimental results, the model after adding Soft NMS can improve the confidence of overlapping objects. As shown in Figure 7, the confidence of overlapping objects Gun and Wrench are both improved.

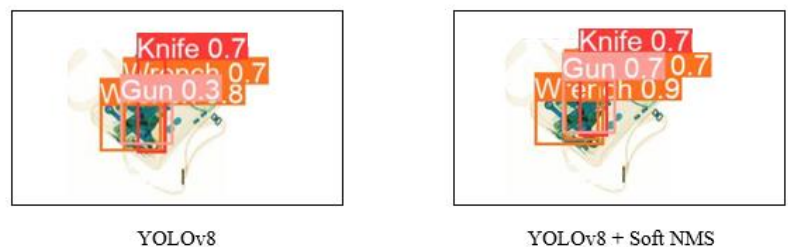

Figure 7: Test result comparison chart

## 2.4 Mosaic

The Mosaic data enhancement method was proposed in the YOLOv4 paper [11]. The main idea is to randomly crop four pictures and stitch them into one picture as training data. The advantage of this is that the background of the picture is enriched, and the four pictures are stitched together, which increases the batch size in disguise. When performing batch normalization, four pictures are also calculated, so the dependence on the batch size itself is not very large, and a single GPU can use more data for training. At the same time, this can also enrich the background of the detected object. In the standardized BN calculation, the data of four pictures will be calculated at a time.

The implementation method of Mosaic is as follows. 1. Stitch the four pictures. 2. Each picture has its corresponding frame, four pictures form a new picture, and the corresponding frame of this picture is obtained at the same time. 3. Passing new pictures into the neural network for learning is equivalent to passing in four original pictures for learning. As shown in Figure 8.

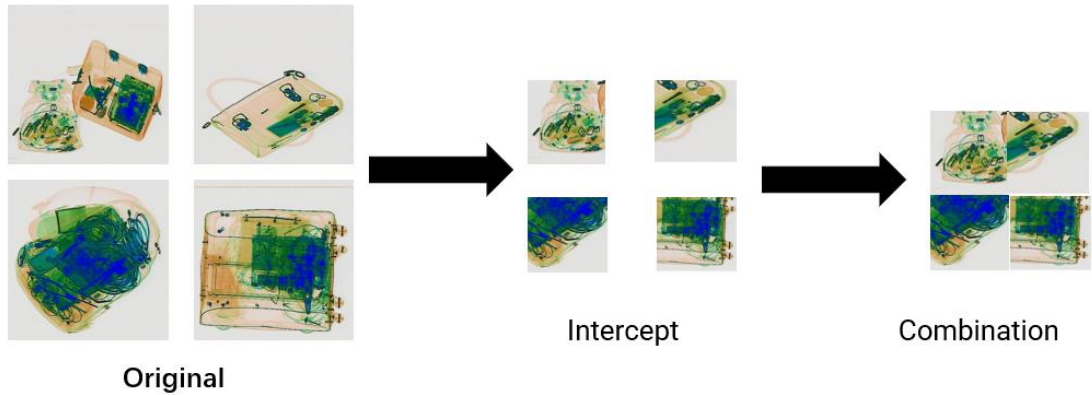

Figure 8: Image Processed by Mosaic

## 3. DATA SET

### 3.1 Dataset

Regarding the datasets, to the best of our knowledge, there are only three published X-ray benchmarks, namely GDXray[22], OPIXray[6] and SIXray[5]. Both GDXray and SIXray are built for object detection tasks, and OPIXray's images are synthetic. We make a detailed comparison in Table I. After comparison, we decided to use the SIXray dataset for research.

Table 1: Comparison of existing open-source X-ray datasets.

| Dataset | Year | Category | $Np$ | Annotation | | | Color | Task | Data Source |
|---|---|---|---|---|---|---|---|---|---|
| | | | | mixture | Number | Professional | | | |
| **GDXray [29]** | 2015 | 3 | 8150 | × | 8150 | × | Grayscale | Detection | Unknown |
| **OPIXray [10]** | 2020 | 5 | 8885 | √ | 8885 | × | RGB | Detection | Artificial Synthesis |
| **SIXray [9]** | 2019 | 6 | 8929 | × | 1,059,231 | √ | RGB | Detection | Subway Station |

Note: Np (Number of images of prohibited items)

This paper conducts model training based on the SIXray dataset, which was constructed by the Pattern Recognition and Intelligent System Development Laboratory of the University of Chinese Academy of Sciences and consists of 1,059,231 X-ray images. Among them, 8929 contraband items in 5 categories are marked. The content of this data set is relatively complex and challenging. The contraband in the data set includes six categories of guns, knives, wrenches, scissors, hammers, and pliers.

But in the actual detection, the image of SIXray is still too simple. X-ray images [16-17] are characterized by a lack of strong discrimination and contain a lot of noise. In response to the above deficiencies, we collected and compiled a high-quality X-ray image contraband detection set through professional data websites such as Kaggle and Roboflow, named QXray data set. After searching the literature, the contraband that needs to be detected often has two problems of small size and overlapping with other objects. Therefore, this data set takes small targets and overlapping objects as the main search direction. At the same time, the distribution ratio of positive and negative samples is set to 1:100, which often occurs, to simulate the real situation to the greatest extent. For some pictures with unreasonable labels, we use the labelImg program to re-label the objects, and update the modified coordinates to the corresponding files. As shown in Figure 9.Therefore, the data we collect has a high degree of similarity with real-world data in terms of category, quantity, location. The quantity of each contraband is shown in Figure 10. Figure 11 is a partial picture of the QXary dataset.

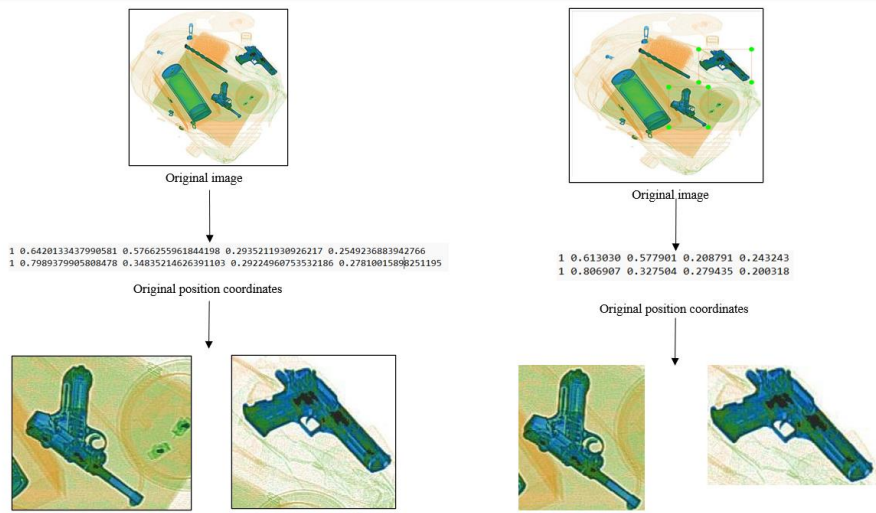

Figure 9: Secondary Annotation

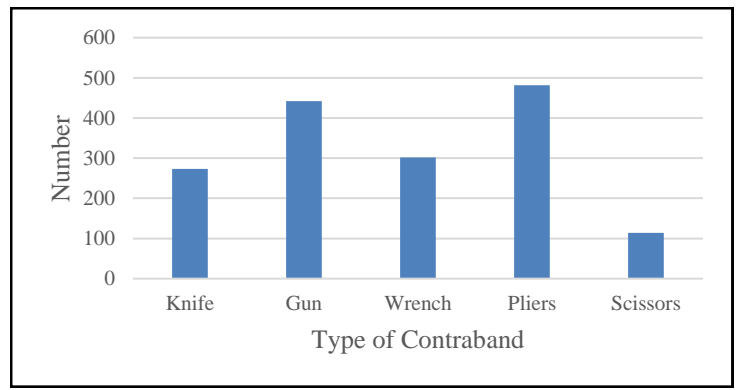

Figure 10: Types of Prohibited Products and Their Quantities

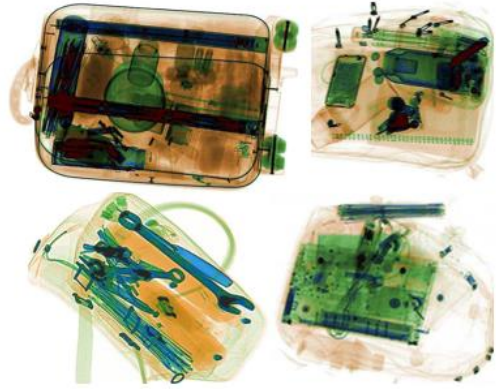

Figure 11:   Partial pictures of QXray dataset

## 4. EVALUATION INDEX

In this paper, we use Precision-Recall Curve, and F1-score to determine whether the performance of the model meets our expected standards. And we select the mean average precision (mAP) of the target detection model to evaluate the accuracy of the model and uses the frames per second (fps) to evaluate the detection efficiency.

### 4.1 Precision-Recall Curve

Precision-Recall Curve (P-R Curve)[18] is a curve with Recall as the x-axis and Precision as the y-axis. The Recall (R) and Precision (P) are calculated according to the equations 5.

$$Precision = \frac{T_P}{T_P + F_P}, \ Recall = \frac{T_P}{T_P + F_N} \qquad (5)$$

where True Positive (TP) means the prediction result is positive class and is judged to be true; False Positive (FP) means the prediction result is positive class but is judged to be false; False Negative (FN) means the prediction result is negative class but is judged to be false.
In object detection, the curve area of P-R Curve can intuitively reflect the performance of the model, so we choose P-R Curve as the evaluation index for the experiment.

### 4.2 F1-score

F1-score is a common indicator for evaluating the accuracy of an algorithm. It comprehensively considers Precision and Recall reflecting the performance of the algorithm in a balanced manner. The equation of F-score is shown in equation 6. where S represents score, P represents precision, and R represents recall.

$$F - score = \frac{(1 + \beta^2) P \times R}{(\beta^2 P) + R} \qquad (6)$$

In general, the weights of Precision and Recall in F1-score are the same when β=1, representing the harmonic mean of Precision and Recall, the equation 7 of which is shown below:

$$F_1 = \frac{2PR}{P + R} = \frac{2T_P}{2T_P + F_P + F_N} \qquad (7)$$

In image detection tasks, F1-score is often used to evaluate the performance of algorithms in object detection tasks. Therefore, we chose F1-score as the asnalysis object for our model evaluation.

### 4.3 Mean Average Precision (mAP)

The calculation of mAP is shown in formula (8).

$$mAP = \int_0^1 P(R)dR \qquad (8)$$

Among them, the calculation method of precision rate P and the calculation method of recall rate R is shown in formula (9), formula (10). TP refers to the number of positive samples that are correctly identified. FP refers to the number of false negative samples. TN refers to the number of negative samples that are correctly identified. FN is the number of false negative positive samples.

$$p = TP/(TP+FP) \qquad (9)$$
$$R = TP/(TP+FN) \qquad (10)$$

MAP is the most used evaluation indicator in object detection, so we use it as an indicator for model comparison. The higher its percentage, the better the model's ability.

### 4.4 Frames Per Second(FPS)

FPS is the frame rate detected per second, used to measure the efficiency of detection. The formula for FPS is shown in formula (11). In this formula, Np represents the number of detect pictures, Ti represents the time taken to detect images.

$$FPS = \frac{Np}{Ti} \qquad (11)$$

This indicator is extremely important for detecting prohibited items. To ensure people's travel efficiency, the speed of detection often needs to be consistent with the speed of the conveyor belt being detected. Therefore, we use it as our evaluation indicator.

## 5. EXPERIMENTAL RESULTS AND ANALYSIS

### 5.1 Experimental Configuration

The experimental environment is based on Python 3.7.0 and implemented using the PyTorch1.8.1 framework in the Windows 11 environment. We recommend readers to use Python 3.7 or higher and PyTorch 1.7 or higher for training. The processor is AMD Ryzen 7 5800H with Radeon Graphics, and the graphics card is NVIDIA GeForce RTX 3060 Laptop GPU for training and testing. In the preprocessing of the image, we set the height and width of the image to 512×512, and the batchsize to 16. During the training process, we found that the model tends to be stable at the 166th epoch. To save computing resources, we changed the number of epochs for all model training to 200.

### 5.2 Result Analysis

To verify the effectiveness of the contraband detection model, the experiments were trained and tested on the SIXray dataset. Compared with the original YOLOv8s, the mAP0.5 is only improved by 0.2%, but the average mAP 0.5:0.9 is increased by 2.5%. This shows that the model's ability to judge blurry objects has been improved. For some objects with low confidence, it is also possible to achieve greater confidence in identifying them.

At the same time, the mAP of each type of detected object is increasing. This indicates that our modifications are effective and can improve YOLOv8's ability to detect prohibited items.

According to data query [19], the main security inspection equipment in my country is the X-ray machine. When the passenger puts the goods on the conveyor belt, the pictures taken are transmitted to the host computer at a speed of 25 frames per second for contraband detection. We ended up with a transfer speed of 40.98, which handled the test site's peak demand well.

Table 2: Model Performance

| Models | Gun | Knife | Wrench | Pliers | Scissors | mAP_0.5(%) | mAP_0.5:0.9(%) | FPS |
|---|---|---|---|---|---|---|---|---|
| YOLOV8s | 90.8 | 97.3 | 91.5 | 92 | 87.9 | 92.4 | 69.8 | 41.83 |
| CSS-YOLO | 91.5 | 97.9 | 91.7 | 93 | 89.5 | 92.6 | 72.3 | 40.98 |

Figure 12 represents the mAP fluctuations during the training process. It is easy to see that YOLOv8s has a small fluctuation range, while CSS-YOLO has a significant fluctuation. The smaller the fluctuation, the easier it is for the model to reach the critical value, so our model's detection ability will be better. The result also indicates that the modified model parameters are higher, and the effect is better.

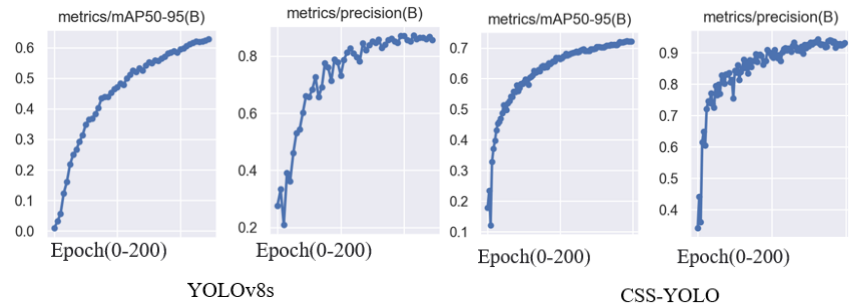

Figure 12: Comparison of experimental results

Figure 13 shows the classification loss of CSS-YOLO and YOLOv8 during training. Due to the large number of iterations and the different gradients of the training loss, we take 20 as an order of magnitude and convert the classification loss to logarithmic data. It can be clearly seen that CSS-YOLO can exceed the performance of YOLOv8 both in terms of loss rate and the result.

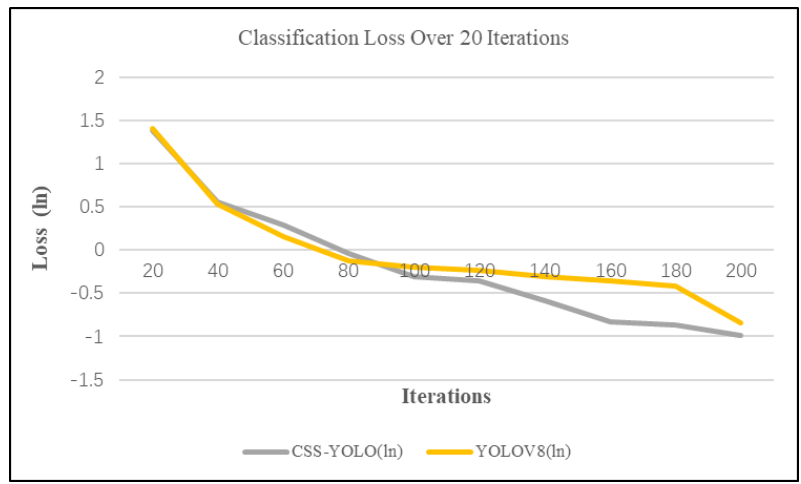

Figure 13: Comparison of the classification loss

Figures 14 and 15 show the experimental results of CSS-YOLO and YOLOv8 on the X-ray dataset. It can be seen from the Precision-Recall Curve (P-R Curve) graph and the F1-Confidence graph that the curve of CSS-YOLO overall class contains a larger area, and the F1 value is larger. This shows that our changes had a positive impact on the model.

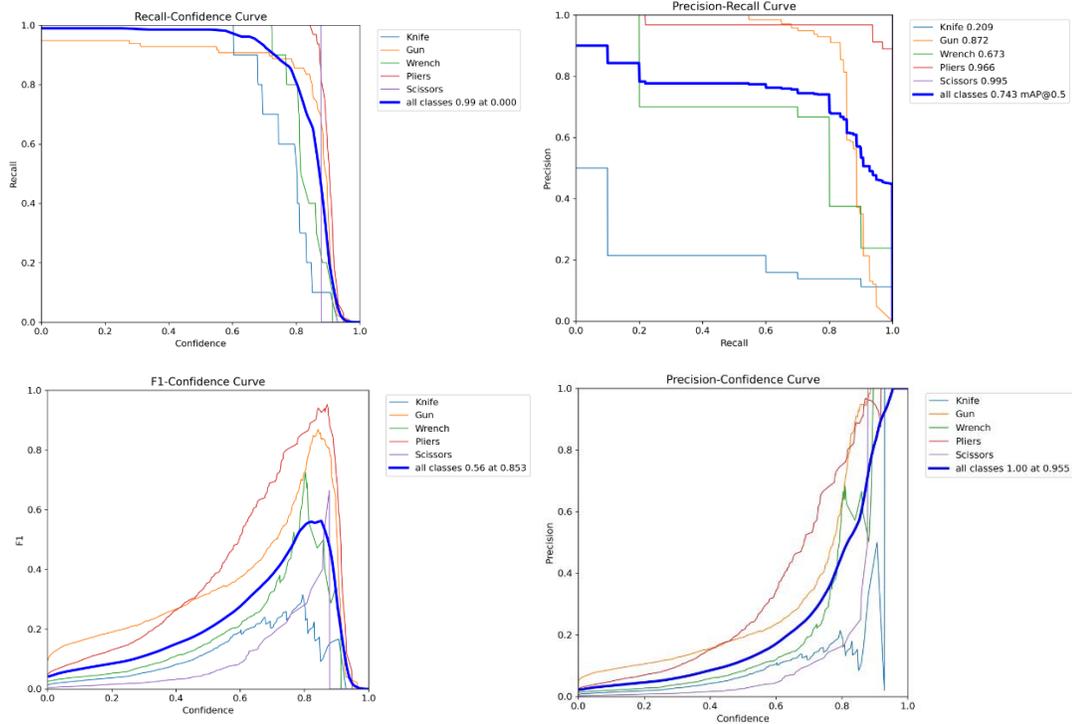

Figure 14: The evaluation index of the YOLOv8s

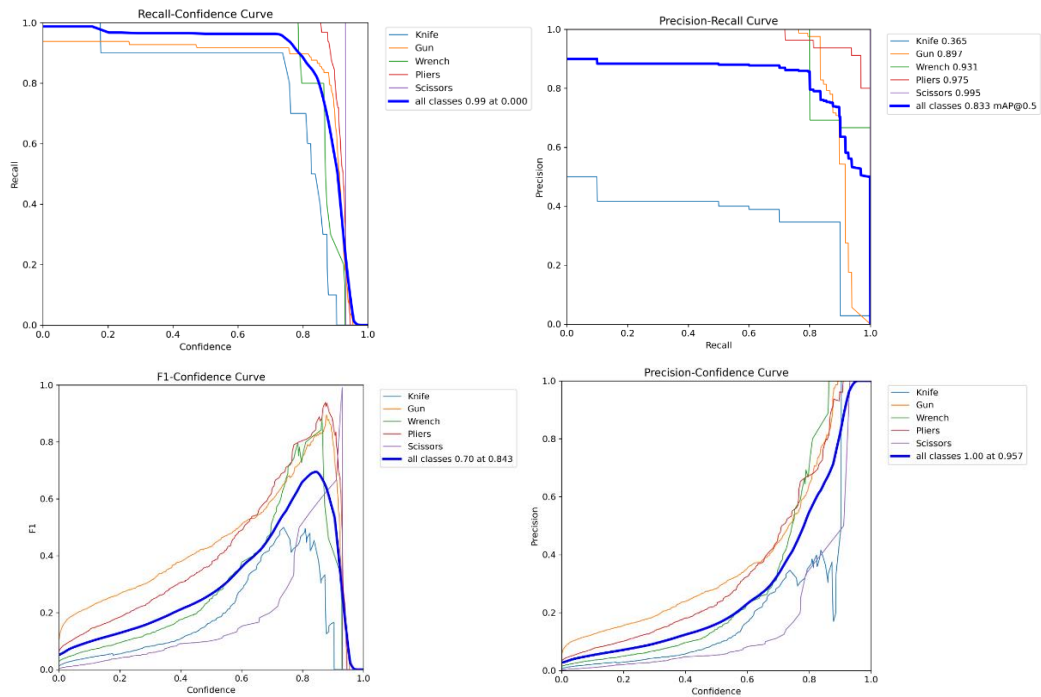

Figure 15: The evaluation index of the CSS-YOLO

Judging from the test results. We can see from Figure 16 that CSS-YOLO can detect overlapping contraband that YOLOv8 cannot detect. From Figure 17 we can see that for small objects such as knives, the detection ability of CSS-YOLO is better than that of YOLOv8. This shows that the modification for overlapping targets and small targets is effective.

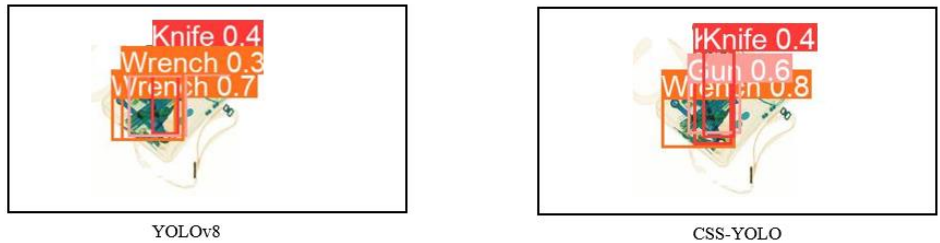

Figure 16: Comparison of detection effects (1)

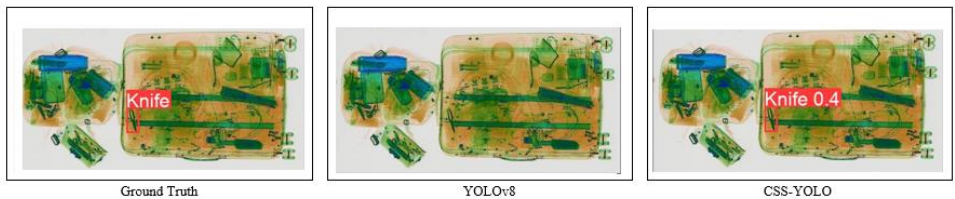

Figure 17: Comparison of detection effects (2)

To verify the advanced nature of the model in this paper, the results are compared with Faster R-CNN[20], Mask R-CNN[21], SSD512[22], DETR[13], YOLOv3[23], YOLOv5s and YOLOX [24], etc. compared with mainstream results. We use the same SIXray dataset to train and test the models. Table 3 shows the comparison of models and the display of evaluation indicators. It can be clearly seen that CSS-YOLO is superior to these models in terms of speed and accuracy.

Table 3: Performance of each model in the dataset

| Models | Gun | Knife | Wrench | Pliers | Scissors | mAP_0.5(%) | mAP_0.5:0.9(%) | FPS |
|---|---|---|---|---|---|---|---|---|
| Faster R-CNN | 90.1 | 80.0 | 79.3 | 58.3 | 88.3 | 84.6 | 49.3 | 11.3 |
| SSD | 88.6 | 72.1 | 63.4 | 76.8 | 82.7 | 76.7 | 45.8 | 35.9 |
| RetinaNet | 88.7 | 76.6 | 79.4 | 86.3 | 88.0 | 83.8 | × | 11.5 |
| ATSS | 88.7 | 78.8 | 80.5 | 87.6 | 89.1 | 84.9 | × | 15.4 |
| AutoAssign | 88.8 | 81.4 | 80.4 | 89.9 | 90.2 | 86.1 | × | 13.3 |
| DETR | 86.6 | 69.2 | 65.1 | 64.3 | 82.4 | 77.3 | 46.3 | 35.7 |
| YOLOv3 | 89.7 | 81.2 | 78.9 | 62.1 | 84.8 | 83.9 | 49.0 | 36.6 |
| YOLOv5s | 88.3 | 84 | 82.1 | 82.0 | 90.6 | 85.8 | 52.9 | 38.8 |
| YOLOX | 89.1 | 86.7 | 85.7 | 90.2 | 93.5 | 89.04 | 57.4 | 37.9 |
| **CSS-YOLO** | **91.5** | **97.3** | **91.7** | **93** | **89.5** | **92.6** | **72.3** | **40.98** |

**Note:** Among them, RetinaNet, ATSS, and AutoAssign come from the data in the paper [53].

### 5.3 Ablation Experiment

To study the specific impact of different improvements on the performance of the model, while simulating the real situation to the greatest extent. Using the original YOLOv8s as the base model, ablation experiments were performed on the self-made dataset QXray. Record the experimental results of the model under each module, and the experimental results are shown in Table IV.

Table 4: Ablation experimental

| Models | mAP_0.5(%) | mAP_0.5:0.9(%) |
|---|---|---|
| Baseline | 70.9 | 60.6 |
| Baseline + CBAM | 78 | 66 |
| Baseline + ST | 72.7 | 64.1 |
| Baseline + Soft NMS | 74.3 | 63.6 |
| Baseline + CBAM + ST + Soft NMS | **83.3** | **72.6** |

It can be seen from Table IV that after adding channel attention and spatial attention to the backbone network of YOLOv8s, the evaluation index mAP of the model is increased by 7.1%, which proves that the CBAM module can improve the contraband detection ability of the model. YOLOv8s+ST increases the ability of the network structure to obtain global features by introducing the Swin Transformer structure. Compared with Baseline, the average mAP of SIXray100 dataset is improved by 3.5%. This article replaces the NMS in the YOLOv8 model with Soft NMS. For overlapping targets, Soft NMS can improve the missed detection rate of overlapping targets, and the average mAP increases by 3%. The final improved model is also better than the single change improvement. The mAP is improved by 12.4%, and the average mAP is improved by 12%.

## 6. CONCLUSIONS

This paper proposes an improvement scheme for contraband detection: 1. For the missed detection problem in contraband detection, we choose to add CBAM to the backbone network, and use the channel attention module and spatial attention module to extract more features information. 2. For the feature extraction problem in small target detection, we introduced the Swin Transformer attention mechanism in the YOLOv8 algorithm to improve the model's ability to acquire features from the global perspective. 3. Solve the problem of overlapping multiple targets in the detection image by introducing Soft NMS to avoid missed detection of targets.
After experiments, CSS-YOLO's mAP0.5:0.9 on the public dataset SIXray reached 72.3%, which is 2.5% higher than YOLOv8. It reached 72.6% on our self-made data set QXray, which is 12% higher than YOLOv8. At the same time, according to the result graph, we can find that the improved YOLOv8 model can more effectively improve the recognition rate of the network for object detection, especially for false detection and missed detection of overlapping blurred objects. The detection speed has also reached 40.95, which can fully meet the detection requirements in real security inspection environments.
In view of the messy detection data in the market, the difficulty of training and detection data does not match and other problems. We took the two directions of small targets and overlapping objects as the entry point and collected 6060 subway security inspection pictures from major data set websites. It aims to provide high-quality data sets for subsequent researchers. At the same time, the method of model deployment is analyzed and summarized. It is hoped that the method proposed in this paper can help managers improve the detection efficiency of contraband and better assist security personnel to complete security inspections. To achieve the purpose of intelligent security inspection and efficient security inspection.
Through the above experiments, we can see that CSS-YOLO has a higher improvement than the traditional model YOLOv8, which proves that our model is usable. In the future, the focus will be on Ultralytics HUB, TensorFlow Lite and PySide for debugging. Build an experimental platform for real-world detection. Optimize the model from the perspective of high precision, light weight, and fast response, and strive to realize the intelligent detection of prohibited items.